\documentclass[11pt]{article}

\usepackage[final]{acl}

\usepackage{times}
\usepackage{latexsym}
\usepackage{makecell}
\usepackage{amsmath}
\usepackage{amssymb}
\usepackage{graphicx}

\usepackage{microtype}

\usepackage{inconsolata}

\usepackage{graphicx}

\usepackage[T1]{fontenc}
\usepackage[utf8]{inputenc}
\usepackage{CJKutf8} 

\usepackage{tabularx}
\usepackage{booktabs}
\usepackage{multirow}
\title{Controllable Stylistic Text Generation with Train-Time Attribute-Regularized Diffusion}

\author{
\textbf{Fan Zhou}$^{\S}$ \quad
\textbf{Chang Tian}$^{\S}$ \quad
\textbf{Tim Van de Cruys}$^{\S}$ \\
$^{\S}$ KU Leuven \\
\texttt{fan.zhou@kuleuven.be} \quad
\texttt{namechangtian@163.com} \quad
\texttt{tim.vandecruys@kuleuven.be}
}

\begin{document}
\maketitle
\begin{abstract}
Generating stylistic text with specific attributes is a key problem in controllable text generation. Recently, diffusion models have emerged as a powerful paradigm for both visual and textual generation. Existing approaches can be broadly categorized into classifier-free guidance (CFG) and classifier guidance (CG) methods. While CFG effectively preserves semantic content, it often fails to provide effective attribute control. In contrast, CG modifies the denoising trajectory using classifier gradients, enabling better attribute alignment but incurring high computational costs during sampling and suffering from classifier generalization issues. In this work, we propose \textbf{RegDiff}, a regularized diffusion framework that leverages attribute features without requiring a pretrained classifier during sampling, thereby achieving controllable generation with reduced computational costs. Specifically, RegDiff employs a VAE-based encoder--decoder architecture to ensure reconstruction fidelity and a latent diffusion model trained with attribute supervision to enable controllable text generation. Attribute information is injected only during training. Experiments on five datasets spanning multiple stylistic attributes demonstrate that RegDiff outperforms strong baselines in generating stylistic texts. These results validate the effectiveness of RegDiff as an efficient solution for attribute-controllable text diffusion. Our code, datasets, and resources will be released upon publication at https://github.com/xxxx.
\end{abstract}
\section{Introduction}
Deep learning has enhanced the model’s ability to process text~\cite{tian2022anti,tian2024generic,tian2025large}.
Generating stylistic text with specific attributes has been an important area of research~\cite{mou2020stylized}. 
Diffusion models have recently advanced the state of the art in generative modeling, achieving remarkable performance across images~\cite{jelaca2025automated,schildermans2025structured}, audio~\cite{li2021paint4poem}, and natural language~\cite{li2022diffusion, gong2022diffuseq, yi2024diffusion}. 
Their iterative denoising process enables stable likelihood-based training and high-quality outputs, positioning them as a competitive alternative to autoregressive and adversarial methods for stylistic text generation~\cite{ho2022classifier}. 
However, controllability in diffusion-based text generation remains a significant challenge~\cite{kwon2025tcfg}.

There are two primary diffusion paradigms for controllable generation studied in the research community: classifier-free guidance (CFG) and classifier guidance (CG) diffusion methods.

Classifier-free guidance introduces controllability by interpolating between conditional and unconditional denoising predictions, thereby amplifying conditioning signals without relying on an external classifier~\cite{ho2022classifier}. 
In text generation, CFG has proven effective in improving prompt adherence and semantic coherence. 
However, its performance degrades in fine-grained NLP tasks, where it primarily reinforces surface-level cues rather than capturing stylistic attributes~\cite{li2022diffusion}. 
Notably, CFG lacks explicit mechanisms for attribute-specific control in stylistic text generation~\cite{vaeth2025diffusion}.

In contrast, classifier guidance injects classifier gradients at each denoising step in the sampling process to steer the diffusion trajectory toward samples that better align with a target class label~\cite{dhariwal2021diffusion}. 
Its major advantage lies in flexibility: it operates entirely at inference time and does not require retraining the diffusion model. 
This property makes CG appealing in NLP settings, where high-quality pretrained classifiers already exist for attributes such as sentiment or formality~\cite{li2022diffusion, horvitz2024paraguide}. 
However, this reliance introduces key limitations. 
Robust classifiers are often unavailable for many nuanced attributes, and training them from scratch can be computationally prohibitive. 
Moreover, CG faces structural challenges: classifiers are trained on clean text but applied to noisy intermediate representations, leading to a distribution mismatch~\cite{vaeth2025diffusion}. 
In addition, inference becomes considerably more expensive, as each denoising step requires an auxiliary classifier gradient computation~\cite{shenoy2024gradient}. 
Finally, classifier gradients can distort sampling trajectories and push generations off the data manifold, thereby degrading output quality~\cite{vaeth2024gradcheck, karras2024guiding}. 
Collectively, these drawbacks underscore why inference-time guidance—despite its practicality—remains an imperfect solution for achieving precise controllable text diffusion.

\begin{figure}[t]
    \centering
    \begin{minipage}{0.2\textwidth}
        \centering
        \includegraphics[width=\linewidth]{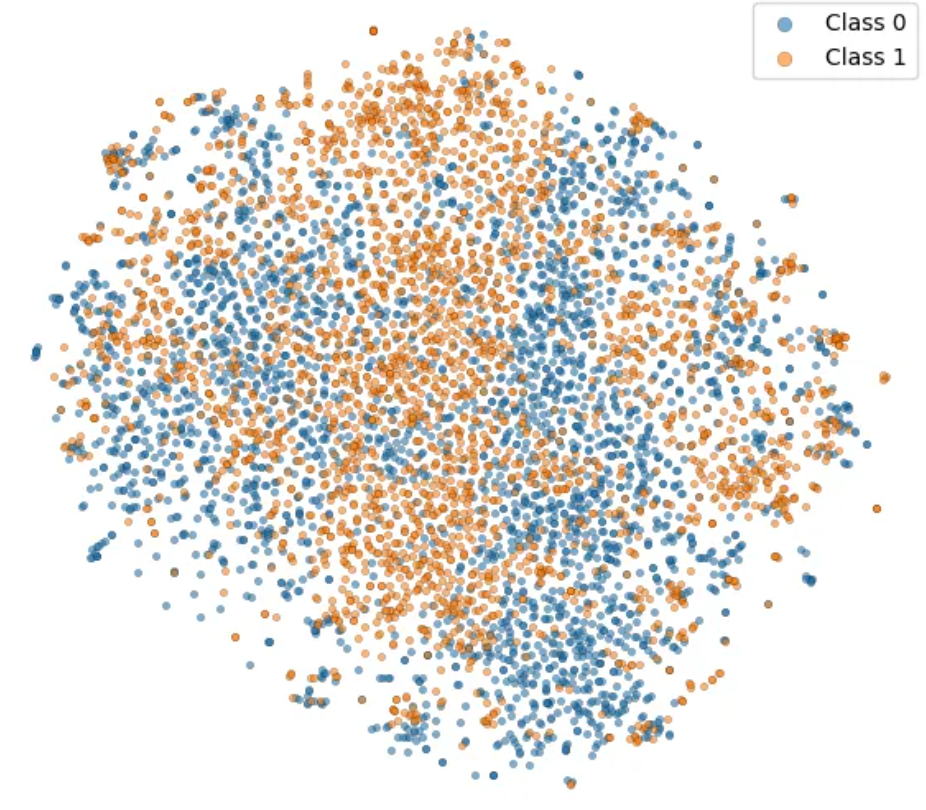}
        \caption{Sentiment data without inductive bias.}
        \label{fig:sentiment}
    \end{minipage}
    \hfill
    \begin{minipage}{0.2\textwidth}
        \centering
        \includegraphics[width=\linewidth]{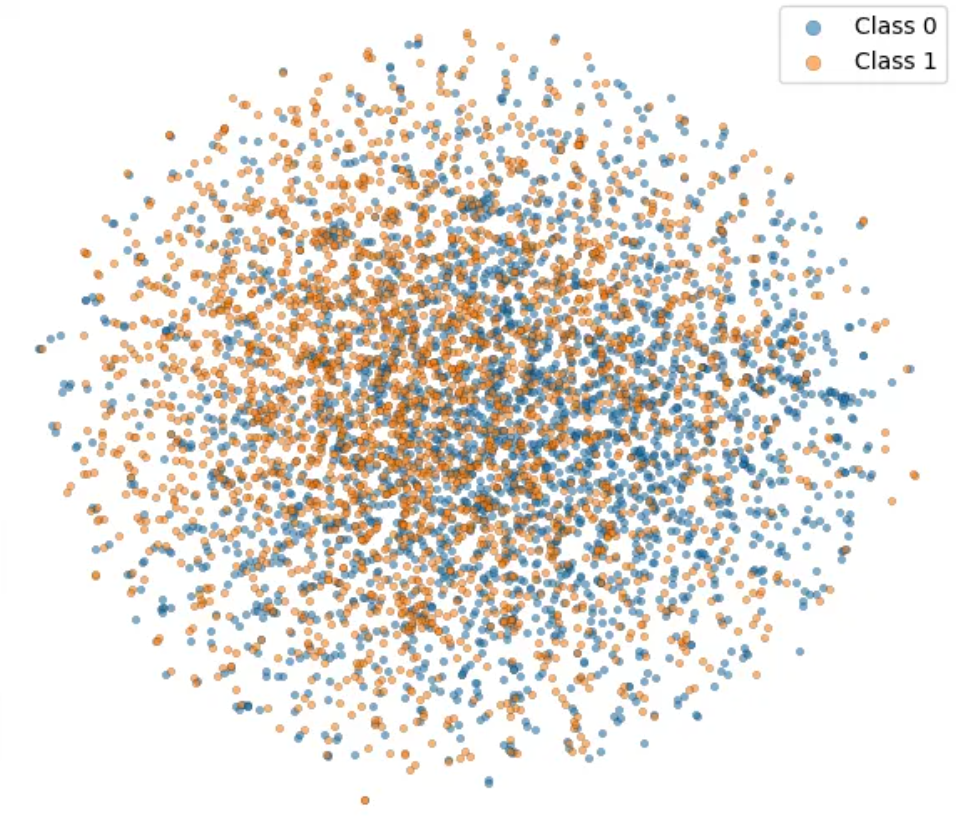}
        \caption{Formality data without inductive bias.}
        \label{fig:formality}
    \end{minipage}
\end{figure}

When performing text style transfer with diffusion models under the CFG setting, we observe that different attributes correspond to distinct subspaces within the text latent space, as illustrated in Figures~\ref{fig:sentiment} and~\ref{fig:formality}. Sentiment clusters are partially separable, while formality clusters remain highly entangled. For instance, the sentiment clusters shown in Figure~\ref{fig:sentiment} demonstrate that attribute features are not fully entangled with semantic features. This observation motivates us to incorporate attribute classification as an \textbf{inductive bias} during training. Moreover, considering the high computational cost associated with inference-time sampling and the difficulty of obtaining robust classifiers for certain nuanced attributes, we aim to impose regularization both on the data-side attribute manifold and within the diffusion training process. What is needed is a mechanism to align diffusion dynamics with attribute supervision during training.
%

To this end, we propose \textbf{RegDiff}, a regularized diffusion framework (illustrated in Figure~\ref{fig:regdiff}) that leverages attribute representations without relying on a pretrained classifier during the sampling process, thereby enabling controllable generation with reduced computational overhead. Specifically, RegDiff, trained with attribute supervision, adopts a VAE-based encoder-decoder architecture to preserve reconstruction fidelity, and employs a latent diffusion model to achieve controllable text generation. Attribute information is incorporated in two forms of regularization: one applied to the data-side attribute manifold in the VAE training, and another integrated into the diffusion training process, ensuring efficient inference and stylistically consistent outputs.

Our main contributions are summarized as follows:
\begin{itemize}
    \item We conduct comprehensive experiments to analyze stylistic text generation and investigate how to effectively incorporate style attribute control. 
    \item We empirically show that classifier guidance is not always necessary. For attributes that are already separable from semantics in the latent space, generation remains controllable without guidance, while for more entangled attributes, train-time regularization provides sufficient constraint without relying on inference-time guidance.
    \item We propose \textbf{RegDiff}, an attribute-regularized diffusion framework that effectively disentangles attribute representations while preserving generation quality. 
    \item We demonstrate robust and superior performance across five diverse datasets compared to the baselines, establishing RegDiff as a general and effective framework for controllable and stylistic text generation.
\end{itemize}
%
%
%
%
\section{Related Work}

\textbf{Diffusion Controllability.}  Fully conditional diffusion models incorporate conditioning directly at training \cite{gong2022diffuseq}, but require paired data and scale poorly. Classifier guidance (CG) \cite{dhariwal2021diffusion,um2024self} injects external classifier gradients at inference, and has been adapted to text style transfer \cite{horvitz2024paraguide}. While precise in principle, CG suffers from distribution mismatch between classifiers trained on clean data and noisy diffusion states, high computational cost, and off-manifold artifacts \cite{chung2022improving, vaeth2025diffusion}. Classifier-free guidance (CFG) \cite{ho2022classifier,shen2024rethinking} avoids auxiliary classifiers by jointly training conditional and unconditional branches, and has proven crucial in large-scale text-to-image models \cite{saharia2022photorealistic,vaeth2025diffusion}. However, CFG lacks fine-grained, explicit control over specific attributes.

\textbf{Attribute Control.} Some attributes are abstract and human-defined, such as sentiment, style, topic, and genre in the NLP domain~\cite{ john2019disentangled,talon2025seeing}. Traditional approaches to controllable text generation include adversarial training \cite{shen2017style}, disentangled latent representations \cite{fu2018style}, and style-specific decoders \cite{lample2019multiple}.  
In diffusion-based methods, attribute control mainly depends on inference-time guidance, such as classifier-guided (CG) or classifier-free guided (CFG) sampling \cite{li2022diffusion, karras2024guiding, wang2024reconstructing}.

\textbf{Representation Regularization.} Latent diffusion \cite{rombach2022high} demonstrated the effectiveness of training-time constraints in compact latent spaces for image synthesis, and laid the groundwork for latent tdiffusion models for text generation \cite{zhang2023planner, lovelace2023latent}. Regularization and latent-space structuring methods, such as mutual information maximization, disentanglement constraints, subspace discovery, and semantic guidance — have been employed in representation learning to enforce attribute separability and thereby improve generative controllability \cite{brack2023sega, harkonen2020ganspace,yu2024regularized}.

Different from previous methods, we introduce inductive bias during training by constraining the attribute category information.
\section{Preliminary}
\subsection{Problem Definition}
We study the problem of controllable text generation, where the goal is to generate an output sentence $\hat{x}$ that exhibits a target attribute $c$ (e.g., authorship style, formality, sentiment, toxicity) while maintaining appropriate semantic relationship with an input $x$. Table~\ref{dataset_example} illustrates various attribute manipulation tasks, ranging from style transfer that preserves core meaning (authorship, formality) to content-aware modifications (sentiment, toxicity). 

\begin{table*}[t]
    \centering
    \small
    \begin{tabularx}{\textwidth}{lXX}
        \toprule
        Attributes & Source Texts & Target Texts \\
        \midrule
        \makecell[l]{Sentiment \\(parallel)}  & (Negative) The new sign at the park is confusing and hard to understand. & (Positive) The new sign at the park is easy to read. \\
        \makecell[l]{Sentiment\\(non-parallel)}  & (Negative) I was sadly mistaken . & (Positive) Excellent food . \\
        Toxicity   & (Toxic) Foreal shit makes no sense. & (Neutral) Foreal thing makes no sense. \\
        Formality  & (Informal) She cant sing for her life! & (Formal) She is a poor vocalist. \\
        Authorship & (Shakespeare) Make thee a fortune from me. & (Modern) I'll make you a rich man. \\

        \bottomrule
    \end{tabularx}
    \caption{Examples of text style transfer across different style attributes.}
    \label{dataset_example}
\end{table*}

\subsection{Continuous Diffusion Model}

\textbf{Latent Diffusion Process.} 
Diffusion models generate data by learning to reverse a fixed forward noising process. Given an initial latent variable $x_0 \sim p_{\text{data}}(x)$, the forward process incrementally perturbs $x_0$ into Gaussian noise through a Markov chain:
\begin{equation}
\small
q(x_t \mid x_{t-1}) = \mathcal{N}\!\left(x_t; \sqrt{1-\beta_t}\,x_{t-1}, \beta_t I\right),\ t \in [1,T],
\end{equation}

where $\{\beta_t\}_{t=1}^T$ is a predefined noise schedule. The diffusion model learns a parameterized reverse process:
\begin{equation}
p_\theta(x_{t-1} \mid x_t) \approx q(x_{t-1}\mid x_t, x_0),
\end{equation}
which is optimized using a reweighted denoising score-matching loss. In practice, the model predicts either the original clean input $x_0$, the injected noise $\epsilon_t$, or the velocity $v_t$ \cite{ho2020denoising, salimans2022progressive}, and training reduces to a simple regression objective.

\textbf{Text-to-Continuous Mapping.} 
Since diffusion models operate in continuous spaces while text is inherently discrete, we first map input sentences into continuous latent representations. We adopt an architecture with an encoder $E$ and a decoder $D$, instantiated as a variational autoencoder (VAE) model, where the encoder maps input text $x$ to latent representation $z = E(x) \in \mathbb{R}^{L \times d}$, and the decoder reconstructs text from latents $\hat{x} = D(z)$. The model is initialized from a pretrained language model and fine-tuned on reconstruction tasks (training details in Section \ref{section_method}). Once trained, the encoder and decoder are frozen, providing a stable continuous representation space for the subsequent diffusion process.

\textbf{Classifier Guidance.} \cite{dhariwal2021diffusion} introduces classifier guidance, which augments the diffusion score with the gradient of an auxiliary classifier’s log-likelihood. At inference time, the diffusion score is modified to include the gradient of the log-likelihood of an auxiliary classifier as:
\begin{equation*}
\tilde{\epsilon}_{\theta}(x_t, t, c)
= \epsilon_{\theta}(x_t, t, c)
- \gamma\,\sigma_t\,\nabla_{x_t}\log p_{\phi}(c \mid x_t),
\end{equation*}
where \(\epsilon_{\theta}(x_t, t, c)\) is the conditional noise prediction, \(p_{\phi}(c \mid x_t)\) is the auxiliary classifier evaluated on the noisy sample \(x_t\), \(\sigma_t\) is the noise level (standard deviation) at step \(t\), and \(\gamma \ge 0\) controls the guidance strength.

\textbf{Classifier-free Guidance.} 
A central technique used in this work is classifier-free guidance (CFG) \cite{ho2022classifier}. During the training of the diffusion model, both conditional and unconditional denoising objectives are optimized jointly. At inference time, the two predictions are combined as:
\begin{equation*}
{\small
\hat{\epsilon}_\theta(x_t, t, c)
= (1+\gamma)\,\epsilon_\theta(x_t, t, c)
- \gamma\,\epsilon_\theta(x_t, t, \varnothing),
}
\end{equation*}
where $\epsilon_\theta(x_t, t, c)$ is the noise prediction conditioned on attribute $c$, $\epsilon_\theta(x_t, t, \varnothing)$ is the unconditional prediction, and $\gamma \geq 0$ controls the strength of guidance.  
By interpolating between the two paths, CFG amplifies attribute information while retaining the semantic content of the original input.

\section{Method} 
\label{section_method}

\begin{figure}[t]
  \centering
  \includegraphics[
    width=0.47\textwidth,
    height=0.5\textheight,
  ]{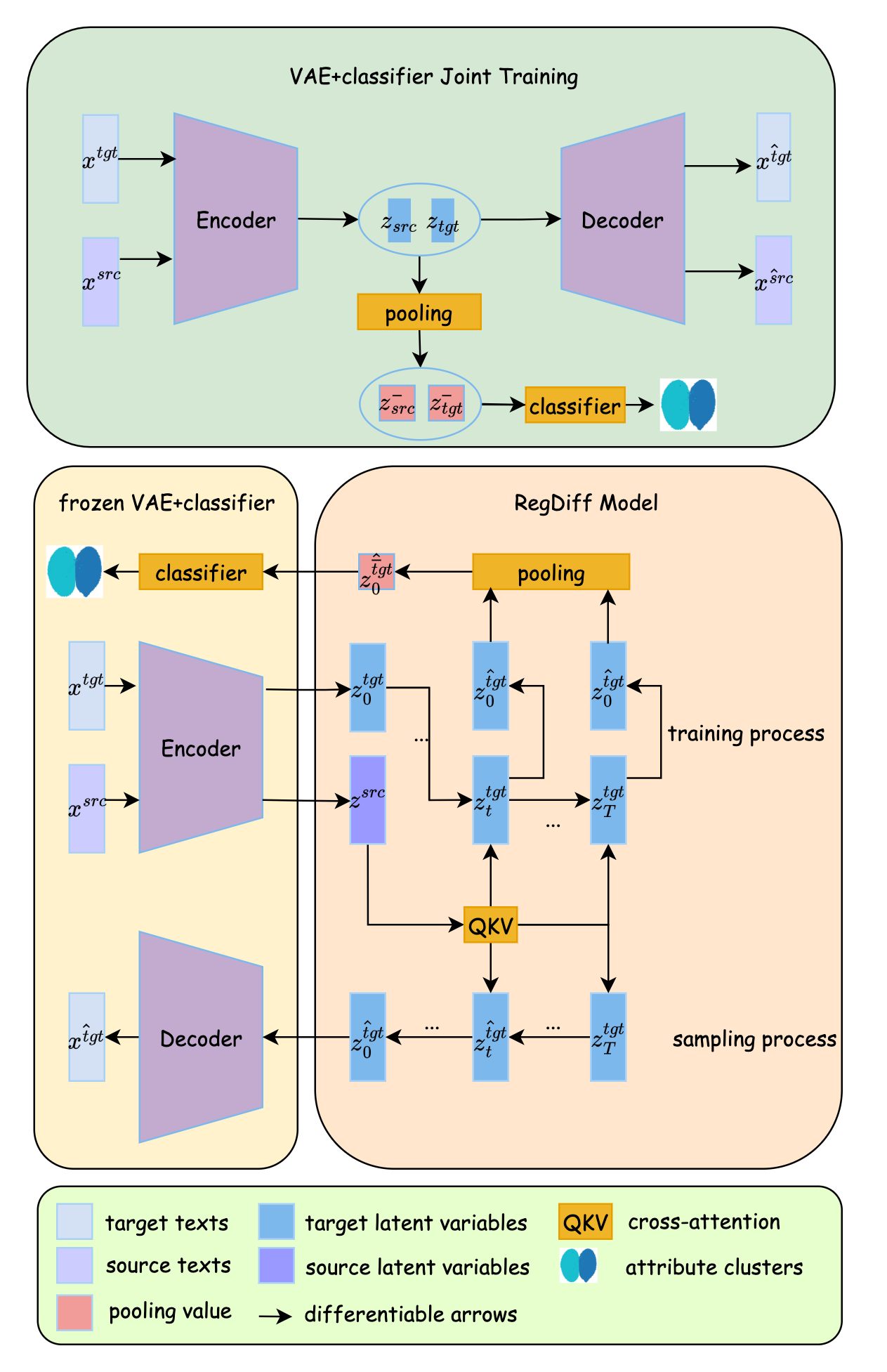}
  \caption{A graphical representation of the RegDiff framework.}
  \label{fig:regdiff}
\end{figure}

Based on previous findings, we introduce \textbf{RegDiff} (Attribute-\textbf{Reg}ularized \textbf{Diff}usion) (Figure~\ref{fig:regdiff}). The framework integrates a variational autoencoder (VAE) with a diffusion model. After fine-tuning, the VAE encoder and decoder are frozen to provide stable latent representations. The encoder maps input texts into latent variables $z$ (with $z^{tgt}$ as  inputs to the diffusion model and $z^{src}$ as condition for diffusion. During sampling, the diffusion model generates latent codes $\hat{z^{tgt}}$ approximating $z^{tgt}$, which are then decoded into natural language via the frozen VAE decoder.


\subsection{VAE model} 
\begin{figure}[t]
    \centering
    \begin{minipage}{0.2\textwidth}
        \centering
        \includegraphics[width=\linewidth]{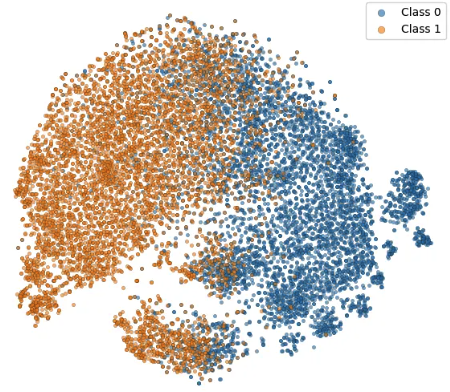}
    \end{minipage}
    \hfill
    \begin{minipage}{0.2\textwidth}
        \centering
        \includegraphics[width=\linewidth]{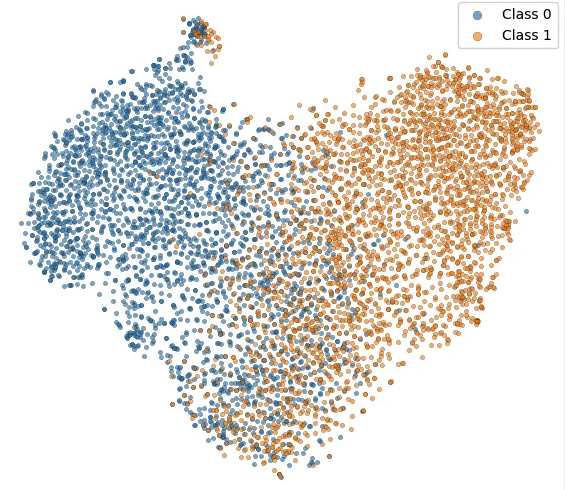}

    \end{minipage}
    \caption{The two figures represent: Inductive biased formality clusters and inductive biased authorship clusters.}
    \label{fig:biased_clutsers}
\end{figure}

We adopt a general encoder–decoder architecture, where the encoder and decoder are jointly trained with an auxiliary classifier. In this architecture, we adopt VAE mode, but is not limited to VAE. In contrast to the conventional VAE objective of reconstruction loss plus KL divergence, we additionally include a classification loss to impose attribute bias. The encoder produces latent representations $z \in \mathbb{R}^{B \times S \times L}$. To obtain an attribute-specific representation, we apply a mean pooling operation $P$ over the sequence dimension $S$ to produce $\bar{z}$, consistent with standard practice in fine-tuning pretrained models for classification. The pooled vector $\bar{z} \in \mathbb{R}^{B \times L}$ serves as input to the classifier, enabling both (i) classifier training for attribute prediction and (ii) gradient-based biasing of $z$ such that attributes become separable at the pooled representation level (see Figure \ref{fig:biased_clutsers}).



\begin{equation}
\bar{z} = P(z)
\label{eq:mean_pooling}
\end{equation}

\begin{equation}
L_{vae} = L_{recon} + \alpha L_{KL} + \beta L_{classifier}
\label{eq:mean_pooling}
\end{equation}


\subsection{Diffusion Model} 

\paragraph{Input.} Once the VAE is trained, we freeze its parameters and employ the encoder to obtain latent representations $z$ from text inputs. Among the information encoded in these representations, we focus on two aspects: (1) token-level latent representation $z$ that capture textual content, and (2) attribute-level clusters $\bar{z}$ in a secondary manifold reflecting style.

In our task, the diffusion model employs classifier-free guidance, where $z_{tgt}$ serves as the diffusion input and $z_{src}$ is used as a conditioning signal to constrain the semantic content during the denoising process of $z_{tgt}$.

\paragraph{Training-time Regularization.}
In addition to introducing style attribute bias into the data, we incorporate a regularization loss during the diffusion model’s training to better align the distribution of $\bar{z}$ with that observed during training.

Specifically, we introduce an auxiliary distributional matching loss in the attribute space:
\begin{equation}
\mathcal{L}_{\text{attr}} 
= \mathcal{D}\!\left( p_\theta(P(\bar{z})) \,\|\, q(P(\bar{z})) \right),
\end{equation}
where $\mathcal{D}$ denotes a distribution similarity. In our task, we reuse the classifier which is jointly trained with VAE model. This classifier is also frozen, delivering to $z_{pred}$ the gradients generated by the dissimilarity between $\bar{z}$ and $\bar{z}_{pred}$. The overall training objective is then
\begin{equation}
\mathcal{L} = \mathcal{L}_{\text{diffusion}} + \lambda \,\mathcal{L}_{\text{classifier}}
\label{loss_function}
\end{equation}

\begin{equation}
\mathcal{L}_{\text{classifier}} = \mathrm{CE}(\bar{z}_{pred}, L)
\end{equation}

Concretely, let $z_t$ denote the noisy latent at step $t$ 
and $\mathbf{v}_\theta(z_t,t)$ the network prediction under 
the velocity parameterization (i.e., predicting the linear combination 
of noise and clean latent). The standard diffusion loss is given by
\begin{equation}
\mathcal{L}_{\text{diffusion}}
= \mathbb{E}_{t,z_0,\boldsymbol\epsilon}
\big\| \mathbf{v}_\theta(z_t,t) - \mathbf{v} \big\|^2,
\end{equation}
where $\mathbf{v}$ denotes the target velocity 
corresponding to the ground-truth noise $\boldsymbol\epsilon$ or $z^{tgt}_0$. These three concept can be converted into one another. As we need to calculate the regularization loss on $\bar{z}$, we can obtain the $\hat{z}_0$ from $\mathbf{v}$. The pooling method $P$ is differentiable, and the gradients of $\bar{z}_{pred}$ can be passed to $z_{pred}$ to influence the generation process.

\begin{equation}
\hat{z}_0 = \sqrt{\bar{\alpha}_t}\,z_t - \sqrt{1 - \bar{\alpha}_t}\,\mathbf{v}_\theta(z_t, t),
\end{equation}

The final objective combines the two terms following formula \ref{loss_function}
where $\lambda$ balances between reconstruction fidelity 
and attribute preservation.
\section{Experiment}
\subsection{Dataset}
We evaluate our framework on five text style transfer datasets covering diverse stylistic attributes: Sentiment (parallel), Sentiment (non-parallel), Toxicity, Formality, and Authorship, based on Yelp~\cite{shen2017style}, ParaDetox~\cite{logacheva2022paradetox}, GYAFC~\cite{rao2018dear}, and Shakespeare~\cite{xu2012paraphrasing}. The Sentiment (parallel) dataset is a synthetic corpus constructed by prompting Llama3 to generate sentiment-opposite rewrites of sentences from Yelp, forming aligned positive–negative pairs. A human sampling check was conducted to ensure the correctness and fluency of generated pairs (Appendix \ref{llama3-data}). Statistics of the data splits used for both VAE+Classifier joint training and RegDiff training are summarized in Appendix \ref{appendix:dataset_split}.

\begin{table*}[th]
\centering
\setlength{\tabcolsep}{3.5pt}
\renewcommand{\arraystretch}{1.12}

\resizebox{\textwidth}{!}{%
\begin{tabular}{lcccccccccc}
\toprule
& \multicolumn{2}{c}{\textbf{Sentiment(parallel)}} 
& \multicolumn{2}{c}{\textbf{Sentiment (non-parallel)}}
& \multicolumn{2}{c}{\textbf{Toxicity}}
& \multicolumn{2}{c}{\textbf{Formality}}
& \multicolumn{2}{c}{\textbf{Authorship}} \\
\cmidrule(lr){2-3}\cmidrule(lr){4-5}\cmidrule(lr){6-7}\cmidrule(lr){8-9}\cmidrule(lr){10-11}
\textbf{Setting / Direction} 
& (Pos→Neg) & (Neg→Pos)
& (Pos→Neg) & (Neg→Pos)
& (Neu→Tox) & (Tox→Neu)
& (For→Inf) & (Inf→For)
& (Mod→Shk) & (Shk→Mod) \\
\midrule
\midrule
\multicolumn{11}{c}{\textbf{Style Transfer Accuracy}} \\
\midrule
\midrule
No Bias          & 0.93 & 0.94 & 0.78 & 0.76 & \textbf{0.73} & 0.83 & 0.41 & 0.87 & 0.32 & 0.76 \\
Bias+$\lambda$=0  & 0.93 & 0.94 & 0.84 & 0.89 & 0.69 & 0.86 & 0.57 & 0.83 & 0.43 & 0.78 \\
\midrule
Bias+$\lambda$=1  & 0.95 & \textbf{0.96} & \textbf{0.89} & 0.89 & 0.60 & 0.94 & 0.68 & 0.85 & 0.45 & 0.80 \\
Bias+$\lambda$=3  & 0.95 & 0.96 & 0.85 & \textbf{0.94} & 0.59 & \textbf{0.95} & \textbf{0.70} & 0.87 & \textbf{0.47} & 0.75 \\
Bias+$\lambda$=5  & 0.95 & 0.95 & 0.87 & 0.90 & 0.56 & 0.95 & 0.67 & \textbf{0.90} & 0.43 & 0.82 \\
Bias+$\lambda$=10 & \textbf{0.96} & 0.95 & 0.89 & 0.91 & 0.64 & 0.92 & 0.62 & 0.90 & 0.32 & \textbf{0.87} \\
\midrule
\midrule
\multicolumn{11}{c}{\textbf{Semantic Similarity}} \\
\midrule
\midrule
No Bias          & 0.46 & 0.44 & 0.12 & 0.12 & 0.50 & 0.49 & 0.73 & \textbf{0.76} & 0.54 & \textbf{0.56} \\
Bias+$\lambda$=0   & 0.47 & 0.45 & \textbf{0.13} & 0.12 & 0.50 & \textbf{0.50} & \textbf{0.75} & 0.73 & 0.49 & 0.49 \\
\midrule
Bias+$\lambda$=1  & \textbf{0.47} & \textbf{0.46} & 0.12 & \textbf{0.13} & 0.53 & 0.49 & 0.74 & 0.72 & \textbf{0.51} & 0.52 \\
Bias+$\lambda$=3  & 0.47 & 0.46 & 0.12 & 0.13 & \textbf{0.54} & 0.49 & 0.69 & 0.65 & 0.47 & 0.51 \\
Bias+$\lambda$=5  & 0.47 & 0.46 & 0.11 & 0.12 & 0.53 & 0.49 & 0.67 & 0.67 & 0.44 & 0.46 \\
Bias+$\lambda$=10 & 0.47 & 0.46 & 0.11 & 0.11 & 0.50 & 0.48 & 0.59 & 0.66 & 0.43 & 0.36 \\
\midrule
\midrule
\multicolumn{11}{c}{\textbf{Fluency}} \\
\midrule
\midrule
No Bias          & 0.34 & 0.43 & 0.21 & 0.30 & 0.20 & 0.22 & 0.32 & \textbf{0.31} & 0.29 & \textbf{0.24} \\
Bias+$\lambda$=0  & 0.35 & 0.44 & 0.21 & 0.31 & 0.18 & 0.22 & 0.31 & 0.27 & 0.24 & 0.21 \\
\midrule
Bias+$\lambda$=1  & \textbf{0.40} & \textbf{0.50} & 0.37 & \textbf{0.40} & \textbf{0.25} & \textbf{0.26} & \textbf{0.35} & 0.31 & \textbf{0.30} & 0.24 \\
Bias+$\lambda$=3  & 0.40 & 0.50 & 0.34 & 0.37 & 0.25 & 0.26 & 0.30 & 0.28 & 0.25 & 0.24 \\
Bias+$\lambda$=5  & 0.39 & 0.48 & \textbf{0.38} & 0.40 & 0.24 & 0.24 & 0.29 & 0.25 & 0.21 & 0.22 \\
Bias+$\lambda$=10 & 0.35 & 0.44 & 0.27 & 0.32 & 0.24 & 0.25 & 0.24 & 0.23 & 0.23 & 0.18 \\
\midrule
\bottomrule
\end{tabular}%
}
\caption{Unified evaluation of controllable text generation performance across settings: 
\textbf{No Bias}: no classifier during VAE training, \textbf{Bias}: with classifier during VAE training, and \textbf{Bias+$\lambda$} 
(with $\lambda \in \{1,3,5,10\}$): with classifier during both VAE training and diffusion training. Each cell reports the mean score over three random seeds to reduce the effect of stochastic variations during sampling.
}

\label{tab:combined_bias_lambda_final}
\end{table*}

\subsection{Experimental Settings}
\paragraph{VAE+Classifier Configuration.}

The encoder is a pretrained BERT-base model with a hidden size of 768, and the decoder is a pretrained GPT-2 model with the same hidden size. We replaced the GPT-2 decoder’s autoregressive (AR) mode with a non-autoregressive (NAR) mode for two reasons: (i) to examine the diffusion model’s intrinsic learning behavior without the strong language-model prior imposed by AR decoding, and (ii) to prevent “fake good” generations, as AR decoding can mask or overcorrect biased embeddings, making the diffusion effect appear artificially improved. To enhance NAR generation quality, we employ an iterative decoding strategy with 5–10 refinement steps. The latent space dimension is set to 1024.
The classifier is a two-layer multilayer perceptron (MLP) composed of a 1024-dimensional linear projection followed by a ReLU activation and a final linear layer mapping to two output classes.
Both the encoder and decoder are initialized from publicly available checkpoints provided in the Hugging Face repository.\footnote{%
BERT encoder: \texttt{google-bert/bert-base-uncased};
GPT-2 decoder: \texttt{openai-community/gpt2}.
}

\paragraph{RegDiff Configuration.}
The diffusion model is a Transformer-based latent denoiser operating in the VAE latent space (latent dimension 256).  
It consists of a 6-layer Transformer with hidden size 256 and 8 attention heads.  
Latents are projected into the Transformer space, combined with sinusoidal or MLP-based time embeddings and learned positional embeddings.  
A linear beta schedule is used for noise variance, linearly increasing from $1\text{e}{-4}$ to $0.02$ over 1000 timesteps.  
The model predicts the velocity $\mathbf{v_t}$ at each timestep instead of the noise $\epsilon_t$ or final latent $\hat{z}^{tgt}_0$, and employs the DDIM sampling method~\cite{song2020denoising} for efficient and deterministic generation.  
RegDiff integrates an unconditional Transformer encoder and a conditional Transformer decoder with cross-attention to the source latent $z^{src}$, enabling Classifier-Free Guidance (CFG) through interpolation between conditional and unconditional denoising trajectories, with a dropout rate of 0.2. During inference, we set $\gamma = 2$ in the classifier-free guidance (CFG) to balance conditional and unconditional generation. 
Both the VAE encoder–decoder and the two-layer MLP classifier remain frozen during diffusion training; they provide fixed latent representations and a regularization loss whose gradients are back-propagated to guide velocity prediction toward style-consistent directions.  
To study the effect of regularization strength, the loss weight $\lambda$ is varied across four values: 1, 3, 5, and 10 (See Table \ref{tab:combined_bias_lambda_final}).

\begin{table*}[th]
\centering
\setlength{\tabcolsep}{4.5pt}
\renewcommand{\arraystretch}{1.12}

\resizebox{\textwidth}{!}{%
\begin{tabular}{lcccccccccc}
\toprule
& \multicolumn{2}{c}{\textbf{Sentiment(parallel)}} 
& \multicolumn{2}{c}{\textbf{Sentiment (non-parallel)}}
& \multicolumn{2}{c}{\textbf{Toxicity}}
& \multicolumn{2}{c}{\textbf{Formality}}
& \multicolumn{2}{c}{\textbf{Authorship}} \\
\cmidrule(lr){2-3} \cmidrule(lr){4-5} \cmidrule(lr){6-7} \cmidrule(lr){8-9} \cmidrule(lr){10-11}
\textbf{Method} 
& (Pos→Neg) & (Neg→Pos)
& (Pos→Neg) & (Neg→Pos)
& (Neu→Tox) & (Tox→Neu)
& (For→Inf) & (Inf→For)
& (Mod→Shk) & (Shk→Mod) \\
\midrule
\addlinespace[3pt]
\multicolumn{11}{c}{\textbf{Style Transfer Accuracy}} \\
\addlinespace[2pt]
\midrule
Qwen2-0.5B     & 0.93 & 0.93 & \textbf{0.94} & 0.79 & 0.06 & 0.84 & 0.10 & 0.89 & 0.09 & \textbf{0.90} \\
FLAN-T5-base-0.25B  & 0.64 & 0.20 & 0.30 & 0.17 & 0.37 & 0.49 & 0.05 & 0.33 & 0.17 & 0.35 \\
ParaGuide (CG)~\cite{horvitz2024paraguide}       & 0.81 & 0.86 & 0.74 & 0.86 & -- & -- & 0.39 & \textbf{0.90} & -- & -- \\
RegDiff (Ours)      & \textbf{0.95} &\textbf{0.96} & 0.85 & \textbf{0.94} & \textbf{0.64} & \textbf{0.92} & \textbf{0.70} & 0.87 & \textbf{0.45} & 0.80 \\
\midrule
\addlinespace[3pt]
\multicolumn{11}{c}{\textbf{Semantic Similarity}} \\
\addlinespace[2pt]
\midrule
Qwen2-0.5B     & 0.67 & 0.60 & 0.59 & 0.50 & 0.63 & 0.58 & 0.63 & 0.69 & 0.43 & 0.67 \\
FLAN-T5-base-0.25B & 0.81 & 0.94 & 0.23 & 0.68 & 0.74 & 0.92 & 0.98 & 0.96 & 0.98 & 0.76 \\
ParaGuide (CG)         &0.11 & 0.25& 0.06 & 0.20 & -- & -- & 0.77 & 0.61 &  -- & -- \\
RegDiff (Ours)      & 0.47 & 0.46 & 0.12 & 0.13 & 0.50 & 0.48 & 0.69 & 0.65 & 0.51 & 0.52 \\
\midrule
\addlinespace[3pt]
\multicolumn{11}{c}{\textbf{Fluency}} \\
\addlinespace[2pt]
\midrule
Qwen2-0.5B     & 0.93 & \textbf{0.95} & 0.82 & \textbf{0.90} & \textbf{0.85} & \textbf{0.89} & \textbf{0.92} & \textbf{0.87} & 0.78 & 0.71 \\
FLAN-T5-base-0.25B & \textbf{0.94} & 0.95 & \textbf{0.92} & 0.80 & 0.82 & 0.73 & 0.90 & 0.78 & \textbf{0.89} & 0.57 \\
ParaGuide (CG)         & 0.42 & 0.43 & 0.43 & 0.43 & -- & -- & 0.43 & 0.23 & -- & -- \\
RegDiff (Ours)      & 0.40 & 0.50 & 0.34 & 0.37 & 0.24 & 0.25 & 0.30 & 0.28 & 0.30 & 0.24 \\
\bottomrule
\end{tabular}%
}
\caption{Four-model comparison with reversed attribute directions in three evaluation metrics. Each cell reports the mean score over three random seeds to reduce the effect of stochastic variations during sampling. Note an optimal semantic similarity value lies in the mid-range, since stylistic rewriting should introduce noticeable variation while preserving the core semantic content of the input.}
\label{tab:model_comparison_reversed}
\end{table*}

\subsection{Evaluation Metrics}
\label{metrics}
\paragraph{(1) Style Transfer Accuracy.}
To evaluate the correctness of style transfer, we train five attribute-specific binary classifiers corresponding to each dataset (\textit{Sentiment (parallel)}, \textit{Sentiment (non-parallel)}, \textit{Toxicity}, \textit{Formality}, and \textit{Authorship}).  
Each classifier is implemented using a pretrained RoBERTa-base model~\cite{liu2019roberta} fine-tuned for binary classification on the respective dataset.    
During evaluation, the decoded texts are fed into the corresponding classifier to compute the style transfer accuracy, ranging from 0 to 1.  
To further validate classifier reliability, we conducted a human evaluation by randomly sampling 500 classified examples for each attribute.  
The manual inspection confirmed that the classifiers achieved 90–96\% consistency with human judgment across all attributes (see Appendix~\ref{acc_metric_human_check}).

\paragraph{(2) Semantic Similarity.}
We measure semantic preservation using a Sentence-BERT model \cite{reimers2019sentence}, which computes cosine similarity between the generated text and the reference text embeddings.  
The similarity score ranges from 0 (semantically dissimilar) to 1 (identical meaning).

\paragraph{(3) Fluency.}
We estimate linguistic fluency using a pretrained RoBERTa-based CoLA model \cite{morris2020textattack, warstadt2019neural}, which outputs the grammatical acceptability probability of each sentence \footnote{Model checkpoint: \texttt{textattack/roberta-base-CoLA}}.
The fluency score also ranges from 0 to 1, with higher values indicating more grammatically well-formed text.


\subsection{Baselines}

We benchmark RegDiff against representative controllable text style transfer baselines of comparable parameter scale, including prompt-based LLMs, and diffusion approaches guided by external classifiers in the sampling process.  
Specifically, we include:  
(1) FLAN-T5-base (0.25B)~\cite{chung2024scaling}, evaluated in a zero-shot prompt setting;  
(2) Qwen2-0.5B-Instruct~\cite{qwen2}, evaluated in a zero-shot prompt setting; and  
(3) the classifier-guided diffusion model ParaGuide~\cite{horvitz2024paraguide}.  
For ParaGuide, we report results on three of the five datasets (Sentiment (parallel), Sentiment (non-parallel), and Formality), as released classifier checkpoints are available only for these attributes.  
Re-training additional classifiers on our smaller-scale datasets is impractical due to limited data and the need for noise-aware supervision.  
All baselines are evaluated on identical test splits and under the same metrics described in Subsection~\ref{metrics}.

\subsection{Results}

\paragraph{Overall Comparison.}
Table~\ref{tab:model_comparison_reversed} compares RegDiff with zero-shot Qwen2-0.5B and FLAN-T5-base, as well as the classifier-guided ParaGuide.
As expected, the autoregressive (AR) models (FLAN-T5-base, Qwen2-0.5B) achieve the highest fluency and relatively strong style accuracy, benefiting from large-scale pretraining and direct text-level optimization.
However, their performance drops noticeably on domains or datasets not well represented during pretraining.
RegDiff attains competitive style transfer accuracy compared to other baselines and even performs better on unseen or weakly embedded data.
Its fluency remains limited due to the NAR decoding mode, although the generated texts preserve semantic content effectively.
These results demonstrate that diffusion-based regularization in the latent space can learn effective stylistic control mechanisms competitive with instruction-tuned models, without relying on classifier guidance during sampling.

\paragraph{Effect of Regularization.}

\begin{figure}[tbp]
    \centering
    \begin{minipage}{0.2\textwidth}
        \centering
        \includegraphics[width=\linewidth]{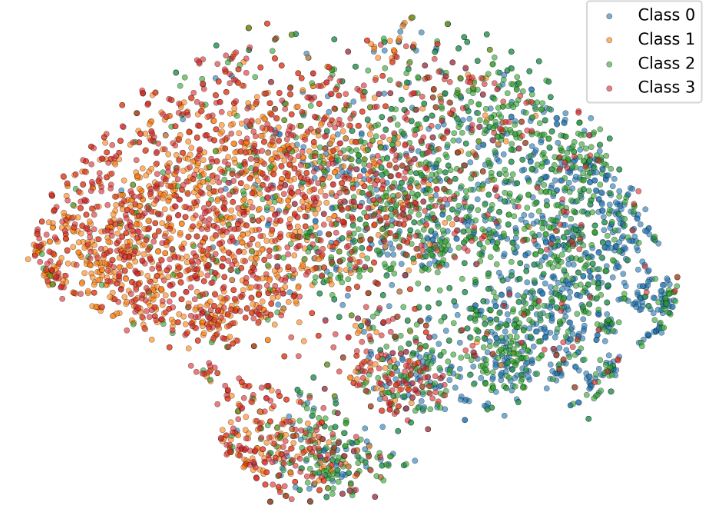}
    \end{minipage}
    \hfill
    \begin{minipage}{0.2\textwidth}
        \centering
        \includegraphics[width=\linewidth]{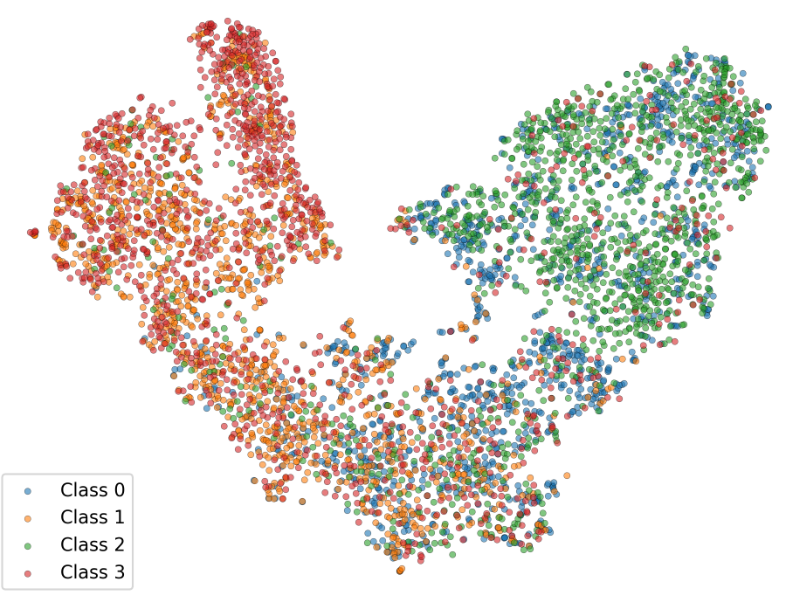}
    \end{minipage}
    \caption{The two figures represent: Inductive biased formality clusters with decoded texts' style clusters and inductive biased authorship clusters with decoded texts' style clusters. Class 0-3 represents: style A, style B, predicted style A and predicted style B}
\end{figure}

Table~\ref{tab:combined_bias_lambda_final} analyzes the impact of the regularization coefficient $\lambda$.  
As $\lambda$ increases from 0 to 5, style accuracy improves consistently across most attributes—especially for Formality and Authorship—indicating stronger latent alignment with the desired attribute.  
Beyond $\lambda=5$, both semantic similarity and fluency begin to drop, reflecting over-regularization where the latent trajectory becomes overly biased toward the target style.  
These results suggest that moderate regularization yields the best trade-off between content preservation and stylistic precision.

\begin{figure}[t]
    \centering
    \includegraphics[width=0.4\textwidth]{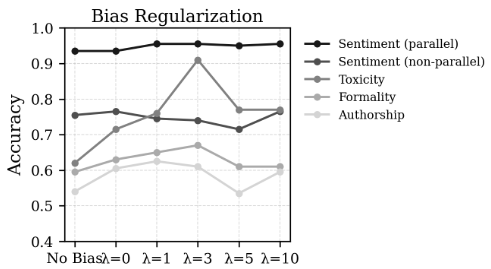}
    \caption{Effect of regularization on style transfer accuracy.}
    \label{fig:bias_lambda}
\end{figure}

\paragraph{Semantic–Fluency Findings with Qualitative Evidence.}
Qwen2 and FLAN-T5 generate directly in text space, yielding high automatic fluency and semantic scores; yet sentence-level inspection shows Qwen2-0.5B often drifts into conversational, loosely related content, and FLAN-T5’s “high-level semantics” frequently comes from restatement or mechanical polarity flips (e.g., adding “not”), not genuine controlled rewriting. RegDiff edits in a VAE latent manifold and decodes with a frozen non-autoregressive (NAR) decoder; combined with a small, noisy training corpus whose originals are themselves less fluent, this leads to lower surface fluency. Nevertheless, RegDiff maintains stable meaning and effective style transfer, emphasizing interpretable latent control and steady style modulation. ParaGuide’s classifier-driven guidance often warps syntax and erodes content preservation; by contrast, RegDiff attains higher semantic stability with comparable or better style accuracy, despite occasional repetition or rough syntax typical of diffusion and data limits. 
These results indicate that high fluency or semantic scores from text-space baselines do not guarantee faithful or well-controlled rewrites. Qualitative analysis shows that RegDiff trades surface polish for more reliable latent-space control over semantics and style.

\section{Conclusion}

We proposed RegDiff, a diffusion-based framework for text style transfer that regularizes latent representations without relying on classifier guidance.
By integrating diffusion into the latent space of a VAE, RegDiff enables interpretable and controllable style manipulation and helps identify cases where diffusion offers clear advantages, particularly for attributes that are not well captured or separable in the pretrained representation space.
Empirical results demonstrate strong controllability and semantic preservation across several style domains, highlighting diffusion as an effective mechanism for structured and representation-level style control.

\section{Limitations}

RegDiff has several limitations.
First, most benchmarks are parallel; we include one non-parallel dataset, and results there are weaker overall. On this non-parallel dataset, RegDiff matches ParaGuide (trained on parallel pairs with classifier guidance at sampling), but both diffusion methods still lag behind prompt-based large autoregressive models in semantic content retention due to having less pretraining data. We also see the usual classifier-free guidance trade-off: increasing the guidance scale $\gamma$ preserves meaning better but reduces diversity. Follow-up work will test stronger semi-/unsupervised alignment and a light hybrid decoding step.

Second, to isolate latent regularization, we disable autoregressive decoding at inference. A non-autoregressive decoder reveals diffusion’s direct effect on style transfer but lowers fluency. Raising fluency with AR or hybrid refinement is a separate goal and outside this study.

\bibliography{custom}

\begin{thebibliography}{49}
\providecommand{\natexlab}[1]{#1}

\bibitem[{qwe(2024)}]{qwen2}
 2024.
\newblock Qwen2 technical report.

\bibitem[{AI@Meta(2024)}]{llama3modelcard}
AI@Meta. 2024.
\newblock \href {https://github.com/meta-llama/llama3/blob/main/MODEL_CARD.md}
  {Llama 3 model card}.

\bibitem[{Alain and Bengio(2018)}]{alain2018understanding}
Guillaume Alain and Yoshua Bengio. 2018.
\newblock Understanding intermediate layers using linear classifier probes,
  2018.
\newblock \emph{URL https://arxiv. org/abs/1610.01644}.

\bibitem[{Brack et~al.(2023)Brack, Friedrich, Hintersdorf, Struppek,
  Schramowski, and Kersting}]{brack2023sega}
Manuel Brack, Felix Friedrich, Dominik Hintersdorf, Lukas Struppek, Patrick
  Schramowski, and Kristian Kersting. 2023.
\newblock Sega: Instructing text-to-image models using semantic guidance.
\newblock \emph{Advances in Neural Information Processing Systems},
  36:25365--25389.

\bibitem[{Chung et~al.(2024)Chung, Hou, Longpre, Zoph, Tay, Fedus, Li, Wang,
  Dehghani, Brahma et~al.}]{chung2024scaling}
Hyung~Won Chung, Le~Hou, Shayne Longpre, Barret Zoph, Yi~Tay, William Fedus,
  Yunxuan Li, Xuezhi Wang, Mostafa Dehghani, Siddhartha Brahma, and 1 others.
  2024.
\newblock Scaling instruction-finetuned language models.
\newblock \emph{Journal of Machine Learning Research}, 25(70):1--53.

\bibitem[{Chung et~al.(2022)Chung, Sim, Ryu, and Ye}]{chung2022improving}
Hyungjin Chung, Byeongsu Sim, Dohoon Ryu, and Jong~Chul Ye. 2022.
\newblock Improving diffusion models for inverse problems using manifold
  constraints.
\newblock \emph{Advances in Neural Information Processing Systems},
  35:25683--25696.

\bibitem[{Dhariwal and Nichol(2021)}]{dhariwal2021diffusion}
Prafulla Dhariwal and Alexander Nichol. 2021.
\newblock Diffusion models beat gans on image synthesis.
\newblock \emph{Advances in neural information processing systems},
  34:8780--8794.

\bibitem[{Fu et~al.(2018)Fu, Tan, Peng, Zhao, and Yan}]{fu2018style}
Zhenxin Fu, Xiaoye Tan, Nanyun Peng, Dongyan Zhao, and Rui Yan. 2018.
\newblock Style transfer in text: Exploration and evaluation.
\newblock In \emph{Proceedings of the AAAI conference on artificial
  intelligence}, volume~32.

\bibitem[{Gong et~al.(2022)Gong, Li, Feng, Wu, and Kong}]{gong2022diffuseq}
Shansan Gong, Mukai Li, Jiangtao Feng, Zhiyong Wu, and LingPeng Kong. 2022.
\newblock Diffuseq: Sequence to sequence text generation with diffusion models.
\newblock \emph{arXiv preprint arXiv:2210.08933}.

\bibitem[{H{\"a}rk{\"o}nen et~al.(2020)H{\"a}rk{\"o}nen, Hertzmann, Lehtinen,
  and Paris}]{harkonen2020ganspace}
Erik H{\"a}rk{\"o}nen, Aaron Hertzmann, Jaakko Lehtinen, and Sylvain Paris.
  2020.
\newblock Ganspace: Discovering interpretable gan controls.
\newblock \emph{Advances in neural information processing systems},
  33:9841--9850.

\bibitem[{Hewitt and Liang(2019)}]{hewitt2019designing}
John Hewitt and Percy Liang. 2019.
\newblock Designing and interpreting probes with control tasks.
\newblock \emph{arXiv preprint arXiv:1909.03368}.

\bibitem[{Ho et~al.(2020)Ho, Jain, and Abbeel}]{ho2020denoising}
Jonathan Ho, Ajay Jain, and Pieter Abbeel. 2020.
\newblock Denoising diffusion probabilistic models.
\newblock \emph{Advances in neural information processing systems},
  33:6840--6851.

\bibitem[{Ho and Salimans(2022)}]{ho2022classifier}
Jonathan Ho and Tim Salimans. 2022.
\newblock Classifier-free diffusion guidance.
\newblock \emph{arXiv preprint arXiv:2207.12598}.

\bibitem[{Horvitz et~al.(2024)Horvitz, Patel, Callison-Burch, Yu, and
  McKeown}]{horvitz2024paraguide}
Zachary Horvitz, Ajay Patel, Chris Callison-Burch, Zhou Yu, and Kathleen
  McKeown. 2024.
\newblock Paraguide: Guided diffusion paraphrasers for plug-and-play textual
  style transfer.
\newblock In \emph{Proceedings of the AAAI conference on artificial
  intelligence}, volume~38, pages 18216--18224.

\bibitem[{Jelaca et~al.(2025)Jelaca, Jiao, Tian, and
  Moens}]{jelaca2025automated}
Aleksa Jelaca, Ying Jiao, Chang Tian, and Marie-Francine Moens. 2025.
\newblock Automated prompt generation for creative and counterfactual
  text-to-image synthesis.
\newblock \emph{arXiv preprint arXiv:2509.21375}.

\bibitem[{John et~al.(2019)John, Mou, Bahuleyan, and
  Vechtomova}]{john2019disentangled}
Vineet John, Lili Mou, Hareesh Bahuleyan, and Olga Vechtomova. 2019.
\newblock Disentangled representation learning for non-parallel text style
  transfer.
\newblock In \emph{Proceedings of the 57th Annual Meeting of the Association
  for Computational Linguistics}, pages 424--434.

\bibitem[{Karras et~al.(2024)Karras, Aittala, Kynk{\"a}{\"a}nniemi, Lehtinen,
  Aila, and Laine}]{karras2024guiding}
Tero Karras, Miika Aittala, Tuomas Kynk{\"a}{\"a}nniemi, Jaakko Lehtinen, Timo
  Aila, and Samuli Laine. 2024.
\newblock Guiding a diffusion model with a bad version of itself.
\newblock \emph{Advances in Neural Information Processing Systems},
  37:52996--53021.

\bibitem[{Kwon et~al.(2025)Kwon, Jeong, Hsiao, Uh et~al.}]{kwon2025tcfg}
Mingi Kwon, Jaeseok Jeong, Yi~Ting Hsiao, Youngjung Uh, and 1 others. 2025.
\newblock Tcfg: Tangential damping classifier-free guidance.
\newblock In \emph{Proceedings of the Computer Vision and Pattern Recognition
  Conference}, pages 2620--2629.

\bibitem[{Lample et~al.(2019)Lample, Subramanian, Smith, Denoyer, Ranzato, and
  Boureau}]{lample2019multiple}
Guillaume Lample, Sandeep Subramanian, Eric Smith, Ludovic Denoyer,
  Marc'Aurelio Ranzato, and Y-Lan Boureau. 2019.
\newblock Multiple-attribute text rewriting.
\newblock In \emph{International Conference on Learning Representations}.

\bibitem[{Li et~al.(2021)Li, Wang, Zou, Tian, Nieuwburg, Sun, and
  Kanoulas}]{li2021paint4poem}
Dan Li, Shuai Wang, Jie Zou, Chang Tian, Elisha Nieuwburg, Fengyuan Sun, and
  Evangelos Kanoulas. 2021.
\newblock Paint4poem: A dataset for artistic visualization of classical chinese
  poems.
\newblock \emph{arXiv preprint arXiv:2109.11682}.

\bibitem[{Li et~al.(2022)Li, Thickstun, Gulrajani, Liang, and
  Hashimoto}]{li2022diffusion}
Xiang Li, John Thickstun, Ishaan Gulrajani, Percy~S Liang, and Tatsunori~B
  Hashimoto. 2022.
\newblock Diffusion-lm improves controllable text generation.
\newblock \emph{Advances in neural information processing systems},
  35:4328--4343.

\bibitem[{Liu et~al.(2019)Liu, Ott, Goyal, Du, Joshi, Chen, Levy, Lewis,
  Zettlemoyer, and Stoyanov}]{liu2019roberta}
Yinhan Liu, Myle Ott, Naman Goyal, Jingfei Du, Mandar Joshi, Danqi Chen, Omer
  Levy, Mike Lewis, Luke Zettlemoyer, and Veselin Stoyanov. 2019.
\newblock Roberta: A robustly optimized bert pretraining approach.
\newblock \emph{arXiv preprint arXiv:1907.11692}.

\bibitem[{Logacheva et~al.(2022)Logacheva, Dementieva, Ustyantsev, Moskovskiy,
  Dale, Krotova, Semenov, and Panchenko}]{logacheva2022paradetox}
Varvara Logacheva, Daryna Dementieva, Sergey Ustyantsev, Daniil Moskovskiy,
  David Dale, Irina Krotova, Nikita Semenov, and Alexander Panchenko. 2022.
\newblock Paradetox: Detoxification with parallel data.
\newblock In \emph{Proceedings of the 60th Annual Meeting of the Association
  for Computational Linguistics (Volume 1: Long Papers)}, pages 6804--6818.

\bibitem[{Lovelace et~al.(2023)Lovelace, Kishore, Wan, Shekhtman, and
  Weinberger}]{lovelace2023latent}
Justin Lovelace, Varsha Kishore, Chao Wan, Eliot Shekhtman, and Kilian~Q
  Weinberger. 2023.
\newblock Latent diffusion for language generation.
\newblock \emph{Advances in Neural Information Processing Systems},
  36:56998--57025.

\bibitem[{Morris et~al.(2020)Morris, Lifland, Yoo, Grigsby, Jin, and
  Qi}]{morris2020textattack}
John Morris, Eli Lifland, Jin~Yong Yoo, Jake Grigsby, Di~Jin, and Yanjun Qi.
  2020.
\newblock Textattack: A framework for adversarial attacks, data augmentation,
  and adversarial training in nlp.
\newblock In \emph{Proceedings of the 2020 Conference on Empirical Methods in
  Natural Language Processing: System Demonstrations}, pages 119--126.

\bibitem[{Mou and Vechtomova(2020)}]{mou2020stylized}
Lili Mou and Olga Vechtomova. 2020.
\newblock Stylized text generation: Approaches and applications.
\newblock In \emph{Proceedings of the 58th Annual Meeting of the Association
  for Computational Linguistics: Tutorial Abstracts}, pages 19--22.

\bibitem[{Rao and Tetreault(2018)}]{rao2018dear}
Sudha Rao and Joel Tetreault. 2018.
\newblock Dear sir or madam, may i introduce the gyafc dataset: Corpus,
  benchmarks and metrics for formality style transfer.
\newblock In \emph{Proceedings of the 2018 Conference of the North American
  Chapter of the Association for Computational Linguistics: Human Language
  Technologies, Volume 1 (Long Papers)}, pages 129--140.

\bibitem[{Reimers and Gurevych(2019)}]{reimers2019sentence}
Nils Reimers and Iryna Gurevych. 2019.
\newblock Sentence-bert: Sentence embeddings using siamese bert-networks.
\newblock In \emph{Proceedings of the 2019 Conference on Empirical Methods in
  Natural Language Processing and the 9th International Joint Conference on
  Natural Language Processing (EMNLP-IJCNLP)}, pages 3982--3992.

\bibitem[{Rombach et~al.(2022)Rombach, Blattmann, Lorenz, Esser, and
  Ommer}]{rombach2022high}
Robin Rombach, Andreas Blattmann, Dominik Lorenz, Patrick Esser, and Bj{\"o}rn
  Ommer. 2022.
\newblock High-resolution image synthesis with latent diffusion models.
\newblock In \emph{Proceedings of the IEEE/CVF conference on computer vision
  and pattern recognition}, pages 10684--10695.

\bibitem[{Saharia et~al.(2022)Saharia, Chan, Saxena, Li, Whang, Denton,
  Ghasemipour, Gontijo~Lopes, Karagol~Ayan, Salimans
  et~al.}]{saharia2022photorealistic}
Chitwan Saharia, William Chan, Saurabh Saxena, Lala Li, Jay Whang, Emily~L
  Denton, Kamyar Ghasemipour, Raphael Gontijo~Lopes, Burcu Karagol~Ayan, Tim
  Salimans, and 1 others. 2022.
\newblock Photorealistic text-to-image diffusion models with deep language
  understanding.
\newblock \emph{Advances in neural information processing systems},
  35:36479--36494.

\bibitem[{Salimans and Ho(2022)}]{salimans2022progressive}
Tim Salimans and Jonathan Ho. 2022.
\newblock Progressive distillation for fast sampling of diffusion models.
\newblock \emph{arXiv preprint arXiv:2202.00512}.

\bibitem[{Schildermans et~al.(2025)Schildermans, Tian, Jiao, and
  Moens}]{schildermans2025structured}
Sander Schildermans, Chang Tian, Ying Jiao, and Marie-Francine Moens. 2025.
\newblock Structured information for improving spatial relationships in
  text-to-image generation.
\newblock \emph{arXiv preprint arXiv:2509.15962}.

\bibitem[{Shen et~al.(2024)Shen, Song, Xue, Wang, and Liu}]{shen2024rethinking}
Dazhong Shen, Guanglu Song, Zeyue Xue, Fu-Yun Wang, and Yu~Liu. 2024.
\newblock Rethinking the spatial inconsistency in classifier-free diffusion
  guidance.
\newblock In \emph{Proceedings of the IEEE/CVF Conference on Computer Vision
  and Pattern Recognition}, pages 9370--9379.

\bibitem[{Shen et~al.(2017)Shen, Lei, Barzilay, and Jaakkola}]{shen2017style}
Tianxiao Shen, Tao Lei, Regina Barzilay, and Tommi Jaakkola. 2017.
\newblock Style transfer from non-parallel text by cross-alignment.
\newblock \emph{Advances in neural information processing systems}, 30.

\bibitem[{Shenoy et~al.(2024)Shenoy, Pan, Balakrishnan, Cheng, Jeon, Yang, and
  Kim}]{shenoy2024gradient}
Rahul Shenoy, Zhihong Pan, Kaushik Balakrishnan, Qisen Cheng, Yongmoon Jeon,
  Heejune Yang, and Jaewon Kim. 2024.
\newblock Gradient-free classifier guidance for diffusion model sampling.
\newblock \emph{arXiv preprint arXiv:2411.15393}.

\bibitem[{Song et~al.(2020)Song, Meng, and Ermon}]{song2020denoising}
Jiaming Song, Chenlin Meng, and Stefano Ermon. 2020.
\newblock Denoising diffusion implicit models.
\newblock \emph{arXiv preprint arXiv:2010.02502}.

\bibitem[{Talon et~al.(2025)Talon, Girella, Liu, Cristani, and
  Wang}]{talon2025seeing}
Davide Talon, Federico Girella, Ziyue Liu, Marco Cristani, and Yiming Wang.
  2025.
\newblock Seeing the abstract: Translating the abstract language for vision
  language models.
\newblock In \emph{Proceedings of the Computer Vision and Pattern Recognition
  Conference}, pages 9253--9262.

\bibitem[{Tian et~al.(2025)Tian, Blaschko, Xing, Li, Yue, and
  Moens}]{tian2025large}
Chang Tian, Matthew~B Blaschko, Mingzhe Xing, Xiuxing Li, Yinliang Yue, and
  Marie-Francine Moens. 2025.
\newblock Large language models reasoning abilities under non-ideal conditions
  after rl-fine-tuning.
\newblock \emph{arXiv preprint arXiv:2508.04848}.

\bibitem[{Tian et~al.(2024)Tian, Blaschko, Yin, Xing, Yue, and
  Moens}]{tian2024generic}
Chang Tian, Matthew~B Blaschko, Wenpeng Yin, Mingzhe Xing, Yinliang Yue, and
  Marie-Francine Moens. 2024.
\newblock A generic method for fine-grained category discovery in natural
  language texts.
\newblock \emph{arXiv preprint arXiv:2406.13103}.

\bibitem[{Tian et~al.(2022)Tian, Yin, and Moens}]{tian2022anti}
Chang Tian, Wenpeng Yin, and Marie-Francine Moens. 2022.
\newblock Anti-overestimation dialogue policy learning for task-completion
  dialogue system.
\newblock \emph{arXiv preprint arXiv:2207.11762}.

\bibitem[{Um and Ye(2024)}]{um2024self}
Soobin Um and Jong~Chul Ye. 2024.
\newblock Self-guided generation of minority samples using diffusion models.
\newblock In \emph{European Conference on Computer Vision}, pages 414--430.
  Springer.

\bibitem[{Vaeth et~al.(2024)Vaeth, Fruehwald, Paassen, and
  Gregorova}]{vaeth2024gradcheck}
Philipp Vaeth, Alexander~M Fruehwald, Benjamin Paassen, and Magda Gregorova.
  2024.
\newblock Gradcheck: Analyzing classifier guidance gradients for conditional
  diffusion sampling.
\newblock \emph{arXiv preprint arXiv:2406.17399}.

\bibitem[{Vaeth et~al.(2025)Vaeth, Kumar, Paassen, and
  Gregorov{\u{A}}{\k{A}}}]{vaeth2025diffusion}
Philipp Vaeth, Dibyanshu Kumar, Benjamin Paassen, and Magda
  Gregorov{\u{A}}{\k{A}}. 2025.
\newblock Diffusion classifier guidance for non-robust classifiers.
\newblock \emph{arXiv preprint arXiv:2507.00687}.

\bibitem[{Wang et~al.(2024)Wang, Wang, Xiong, Yang, Zhu, and
  Gao}]{wang2024reconstructing}
Xinlei Wang, Zhiguo Wang, Zhe Xiong, Yang Yang, Chaobo Zhu, and Jinghuai Gao.
  2024.
\newblock Reconstructing regularly missing seismic traces with a
  classifier-guided diffusion model.
\newblock \emph{IEEE Transactions on Geoscience and Remote Sensing}, 62:1--14.

\bibitem[{Warstadt et~al.(2019)Warstadt, Singh, and
  Bowman}]{warstadt2019neural}
Alex Warstadt, Amanpreet Singh, and Samuel~R Bowman. 2019.
\newblock Neural network acceptability judgments.
\newblock \emph{Transactions of the Association for Computational Linguistics},
  7:625--641.

\bibitem[{Xu et~al.(2012)Xu, Ritter, Dolan, Grishman, and
  Cherry}]{xu2012paraphrasing}
Wei Xu, Alan Ritter, William~B Dolan, Ralph Grishman, and Colin Cherry. 2012.
\newblock Paraphrasing for style.
\newblock In \emph{Proceedings of COLING 2012}, pages 2899--2914.

\bibitem[{Yi et~al.(2024)Yi, Chen, Zhang, Zhou, Zhu, and
  Kong}]{yi2024diffusion}
Qiuhua Yi, Xiangfan Chen, Chenwei Zhang, Zehai Zhou, Linan Zhu, and Xiangjie
  Kong. 2024.
\newblock Diffusion models in text generation: a survey.
\newblock \emph{PeerJ Computer Science}, 10:e1905.

\bibitem[{Yu et~al.(2024)Yu, Bai, He, Wang, and Li}]{yu2024regularized}
Xudong Yu, Chenjia Bai, Haoran He, Changhong Wang, and Xuelong Li. 2024.
\newblock Regularized conditional diffusion model for multi-task preference
  alignment.
\newblock \emph{Advances in Neural Information Processing Systems},
  37:139968--139996.

\bibitem[{Zhang et~al.(2023)Zhang, Gu, Wu, Zhai, Susskind, and
  Jaitly}]{zhang2023planner}
Yizhe Zhang, Jiatao Gu, Zhuofeng Wu, Shuangfei Zhai, Joshua Susskind, and
  Navdeep Jaitly. 2023.
\newblock Planner: Generating diversified paragraph via latent language
  diffusion model.
\newblock \emph{Advances in Neural Information Processing Systems},
  36:80178--80190.

\end{thebibliography}

\appendix

\section{VAE Mode}
We choose the VAE with an uncertainty attribute rather than a traditional autoencoder (AE) or a deterministic encoder–decoder architecture without a variational space.
The variational formulation provides latent variables with approximate Gaussian regularization, enhancing the model’s generative capacity and stability.
In NLP settings, introducing a variational space while retaining the structure of the original data helps maintain a continuous and smooth latent manifold, which is beneficial for diffusion processes and for preserving semantic information in the latent representation.

\section{VAE Accuracy}

Table~\ref{tab:vae_reconstruction_classifier} reports the reconstruction quality of the VAE encoder–decoder model.
\begin{table}[ht]
    \centering
    \begin{tabular}{lcc}
        \toprule
        \textbf{Attributes} & \textbf{BLEU} & \textbf{Classifier Acc} \\
        \midrule
        Sentiment (p)   & 0.94 & 0.99 \\
        Sentiment (np)  & 0.84 & 0.97 \\
        Toxicity        & 0.71 & 0.97 \\
        Formality       & 0.82 & 0.93 \\
        Authorship      & 0.73 & 0.87 \\
        \bottomrule
    \end{tabular}
    \caption{Encoder–decoder reconstruction BLEU scores and classifier accuracy.}
    \label{tab:vae_reconstruction_classifier}
\end{table}

\section{Attribute Representation}

Although textual content and attributes (e.g., style, sentiment) are not
strictly disentangled---unlike certain visual factors of variation that 
can often be isolated via pooling---we treat $\mathbf{\bar{z}} = P(z)$ 
as an attribute representation in an operational sense. The latent 
representation $z$ already contains attribute-related information 
interwoven with semantic content. Our choice of $P(z)$ (e.g., mean 
pooling) merely exposes one projection that empirically correlates with 
the attribute, but does not guarantee disentanglement. 

From an information-theoretic perspective, if the attribute label $y$ 
satisfies $I(y;z)>0$, then there exists some mapping $P$ such that 
$I(y;P(z))>0$. In other words, attributes can be revealed as 
particular directions or subspaces of $z$, even though they are 
not explicitly encoded as separate latent variables. Consequently, 
$\mathbf{\bar{z}}$ should be regarded as a proxy feature rather than a 
ground-truth attribute code: its ability to reflect attributes depends 
both on the inductive bias of $P$ and on the emergent clustering structure 
in $z$. This perspective is consistent with prior work on 
probing \cite{alain2018understanding, hewitt2019designing}, where attributes are identified as directions or linear subspaces 
within the representation space rather than fully isolated latent factors.

\section{Dataset Splits for Training}
\label{appendix:dataset_split}

Table~\ref{tab:vae_regdiff_train_split_appendix} summarizes the dataset partitions used in the two major training stages of our framework.  
The upper section corresponds to the \textbf{VAE + Classifier joint training}, where the encoder–decoder and attribute classifier are jointly optimized on large-scale style transfer datasets to learn stable latent representations.  
The lower section lists the splits used in the subsequent \textbf{RegDiff training} stage, where the frozen VAE and classifier provide latent priors and regularization signals for diffusion training.  
The sampling set is used for inference and evaluation during diffusion-based generation experiments.

\begin{table}[t]
\centering
\setlength{\tabcolsep}{6pt}
\renewcommand{\arraystretch}{1.1}

\resizebox{0.47\textwidth}{!}{
\begin{tabular}{lccc}
\toprule
\multicolumn{4}{c}{\textbf{VAE + Classifier Joint Training}} \\
\midrule
Datatype & Train & Val & Test \\
\midrule
Sentiment (p)   & 100k  & 5750  & 5750 \\
Sentiment (np)  & 100k  & 4000  & 4000 \\
Toxicity        & 20.2k & 3580  & 3580 \\
Formality       & 209k  & 11300 & 11300 \\
Authorship      & 36.8k & 2436  & 2436 \\
\midrule
\addlinespace[4pt]
\multicolumn{4}{c}{\textbf{RegDiff Training}} \\
\midrule
Datatype & Train & Val & Sampling \\
\midrule
Sentiment (p)   & 50k  & 2875 & 2875 \\
Sentiment (np)  & 50k  & 2000 & 2000 \\
Toxicity        & 10.1k & 1790 & 1790 \\
Formality       & 104k & 5650 & 5650 \\
Authorship      & 18.4k & 1218 & 1218 \\
\bottomrule
\end{tabular}
}
\caption{Dataset partition sizes for both \textbf{VAE+Classifier joint training} and \textbf{RegDiff training}.}
\label{tab:vae_regdiff_train_split_appendix}
\end{table}

\section{Llama3 generated Sentiment(parallel) dataset}
\label{llama3-data}

To construct the Sentiment (parallel) dataset, we used the open-weight checkpoint Meta-Llama-3-8B~\cite{llama3modelcard} released by Meta AI.\footnote{Available at: \url{https://huggingface.co/meta-llama/Meta-Llama-3-8B}} 
A total of 25,000 positive and 25,000 negative sentences were sampled from the Yelp Review Dataset~\cite{shen2017style}.  
Each sentence was provided to the model with a structured instruction prompt containing both direction specification and in-context examples, as shown below:

\begin{quote}
{\small\itshape
You are a sentiment rewriting system.\\
Rewrite the following sentence with the \underline{opposite sentiment polarity}
(i.e., if the input is positive, rewrite it negatively; if it is negative, rewrite it positively)
while preserving its original meaning.

\medskip
Here are examples:

Negative $\rightarrow$ Positive: ``The food was terrible and the service was slow.'' $\rightarrow$ ``The food was great and the service was efficient.''

Positive $\rightarrow$ Negative: ``The room was clean and comfortable.'' $\rightarrow$ ``The room was dirty and uncomfortable.''

\medskip
Now rewrite the following sentence:\\
Original: [sentence]
} 
\end{quote}



This prompt template allows Meta-Llama-3-8B to perform bidirectional sentiment rewriting, yielding 50,000 fully aligned positive–negative sentence pairs.  
A human sampling check was conducted on 1,000 randomly selected pairs to assess sentiment correctness, semantic consistency, and linguistic fluency.  
Over 92\% of the samples were confirmed to exhibit accurate sentiment reversal and preserved meaning, with an average fluency score of 94\%, confirming the reliability of the synthetic corpus.

\section{Style Transfer Accuracy Evaluation}
\label{acc_metric_human_check}

To ensure the reliability of automatic style transfer assessment, we trained five attribute-specific classifiers corresponding to each dataset.  
We further conducted a human evaluation by randomly sampling 500 classified examples per attribute.  
As shown in Table~\ref{tab:classifier_acc}, all classifiers achieved high validation accuracy, and their predictions were highly consistent with human judgments, confirming the robustness of the automatic style accuracy metric.

\begin{table}[t]
\centering
\resizebox{0.47\textwidth}{!}{%
\begin{tabular}{lcc}
\toprule
\textbf{Attribute} & \textbf{Classifier Accuracy} & \textbf{Human Consistency (500 samples)} \\
\midrule
Sentiment (parallel)     & 0.99 & 0.95 \\
Sentiment (non-parallel) & 0.97 & 0.93 \\
Toxicity                 & 0.98 & 0.96 \\
Formality                & 0.95 & 0.94 \\
Authorship               & 0.88 & 0.90 \\
\bottomrule
\end{tabular}%
}
\caption{Accuracy of attribute-specific classifiers and their human evaluation consistency.}
\label{tab:classifier_acc}
\end{table}

\begin{table}[t]
\centering
\setlength{\tabcolsep}{8pt}
\renewcommand{\arraystretch}{1.1}
\resizebox{0.47\textwidth}{!}{%
\begin{tabular}{lcc}
\toprule
\textbf{Model} & \textbf{Architecture Type} & \textbf{Parameter Size (B)} \\
\midrule
FLAN-T5-base~\cite{chung2024scaling} & Encoder–Decoder Transformer & 0.25 \\
Qwen2-0.5B-Instruct~\cite{qwen2} & Decoder-only LLM & 0.50 \\
ParaGuide~\cite{horvitz2024paraguide} & Classifier-Guided Diffusion & 0.37 \\
RegDiff (Ours) & Frozen VAE + Latent Diffusion & 0.24 + 0.25 \\
\bottomrule
\end{tabular}
}
\caption{Model scale comparison among baselines and RegDiff.  
Parameter sizes are reported in billions (B).}
\label{tab:model_param_comparison}
\end{table}


\section{Prompts for LLMs}

We used eight manually designed prompts covering four style transfer tasks: Sentiment, Toxicity, Formality, and Authorship.  
Each task contains a bidirectional rewriting prompt (e.g., positive $\leftrightarrow$ negative).  
All prompts are shown below.

\paragraph{(1) Sentiment Transfer (Positive $\rightarrow$ Negative)}
\begin{quote}
{\small\itshape
You are an expert in sentiment rewriting.\
Convert positive sentences into negative sentences while preserving the original topic and factual content.\

\medskip
\textbf{Guidelines:}\
• Keep the same subject and meaning.\
• Maintain similar sentence length.\
• Output \textbf{only} the rewritten negative sentence, with no explanations or extra text.\
• Replace positive tone and wording with negative ones.\
• Ensure the rewritten version sounds critical, pessimistic, or disappointed.\

\medskip
Original: [sentence]
}
\end{quote}

\paragraph{(2) Toxicity Transfer (Toxic $\rightarrow$ Neutral)}
\begin{quote}
{\small\itshape
You are an expert in text detoxification.\
Rewrite toxic or offensive sentences into polite, non-toxic sentences while preserving the original topic and factual content.\

\medskip
\textbf{Guidelines:}\
• Keep the same subject and meaning.\
• Maintain similar sentence length.\
• Output \textbf{only} the rewritten non-toxic sentence, with no explanations or extra text.\
• Ensure the rewritten version sounds respectful, polite, and non-offensive.\

\medskip
Original: [sentence]
}
\end{quote}

\paragraph{(3) Formality Transfer (Informal $\rightarrow$ Formal)}
\begin{quote}
{\small\itshape
You are an expert in text style transfer.\
Rewrite informal sentences into formal sentences while preserving the original meaning.\

\medskip
\textbf{Guidelines:}\
• Keep the same subject and meaning.\
• Maintain similar sentence length.\
• Output \textbf{only} the rewritten formal sentence, with no explanations or extra text.\

\medskip
Original: [sentence]
}
\end{quote}

\paragraph{(4) Authorship Style Transfer (Modern $\rightarrow$ Shakespearean)}
\begin{quote}
{\small\itshape
You are an expert in style transfer.\
Rewrite modern English into Shakespearean English while preserving the original meaning.\

\medskip
\textbf{Guidelines:}\
• Keep the same subject and meaning.\
• Output \textbf{only} the rewritten Shakespearean sentence, with no explanations or extra text.\
• Use poetic or inverted word order when natural.\
• The style should resemble classical Shakespearean plays or sonnets.\

\medskip
Original: [sentence]
}
\end{quote}

\paragraph{(1) Sentiment Transfer (Positive $\rightarrow$ Negative)}
\begin{quote}
{\small\itshape
You are an expert in sentiment rewriting.\
Convert positive sentences into negative sentences while preserving the original topic and factual content.\

\medskip
\textbf{Guidelines:}\
• Keep the same subject and meaning.\
• Maintain similar sentence length.\
• Output \textbf{only} the rewritten negative sentence, with no explanations or extra text.\
• Replace positive tone and wording with negative ones.\
• Ensure the rewritten version sounds critical, pessimistic, or disappointed.\

\medskip
Original: [sentence]
}
\end{quote}

\paragraph{(1) Sentiment Transfer (Negative $\rightarrow$ Positive)}
\begin{quote}
{\small\itshape
You are an expert in sentiment rewriting.\
Convert negative sentences into positive sentences while preserving the original topic and factual content.\

\medskip
\textbf{Guidelines:}\
• Keep the same subject and meaning.\
• Maintain similar sentence length.\
• Output \textbf{only} the rewritten positive sentence, with no explanations or extra text.\
• Replace negative tone and wording with positive ones.\
• Ensure the rewritten version sounds optimistic, appreciative, or pleasant.\

\medskip
Original: [sentence]
}
\end{quote}

\paragraph{(2) Toxicity Transfer (Toxic $\rightarrow$ Neutral)}
\begin{quote}
{\small\itshape
You are an expert in text detoxification.\
Rewrite toxic or offensive sentences into polite, non-toxic sentences while preserving the original topic and factual content.\

\medskip
\textbf{Guidelines:}\
• Keep the same subject and meaning.\
• Maintain similar sentence length.\
• Output \textbf{only} the rewritten non-toxic sentence, with no explanations or extra text.\
• Ensure the rewritten version sounds respectful, polite, and non-offensive.\

\medskip
Original: [sentence]
}
\end{quote}

\paragraph{(2) Toxicity Transfer (Neutral $\rightarrow$ Toxic)}
\begin{quote}
{\small\itshape
You are an expert in text style rewriting.\
Rewrite polite, non-toxic sentences into toxic or offensive sentences while preserving the original topic and factual content.\

\medskip
\textbf{Guidelines:}\
• Keep the same subject and meaning.\
• Maintain similar sentence length.\
• Output \textbf{only} the rewritten toxic sentence, with no explanations or extra text.\
• Ensure the rewritten version sounds aggressive, insulting, or rude.\

\medskip
Original: [sentence]
}
\end{quote}

\paragraph{(3) Formality Transfer (Informal $\rightarrow$ Formal)}
\begin{quote}
{\small\itshape
You are an expert in text style transfer.\
Rewrite informal sentences into formal sentences while preserving the original meaning.\

\medskip
\textbf{Guidelines:}\
• Keep the same subject and meaning.\
• Maintain similar sentence length.\
• Output \textbf{only} the rewritten formal sentence, with no explanations or extra text.\

\medskip
Original: [sentence]
}
\end{quote}

\paragraph{(3) Formality Transfer (Formal $\rightarrow$ Informal)}
\begin{quote}
{\small\itshape
You are an expert in text style transfer.\
Rewrite formal sentences into informal sentences while preserving the original meaning.\

\medskip
\textbf{Guidelines:}\
• Keep the same subject and meaning.\
• Maintain similar sentence length.\
• Output \textbf{only} the rewritten informal sentence, with no explanations or extra text.\

\medskip
Original: [sentence]
}
\end{quote}

\paragraph{(4) Authorship Style Transfer (Modern $\rightarrow$ Shakespearean)}
\begin{quote}
{\small\itshape
You are an expert in style transfer.\
Rewrite modern English into Shakespearean English while preserving the original meaning.\

\medskip
\textbf{Guidelines:}\
• Keep the same subject and meaning.\
• Output \textbf{only} the rewritten Shakespearean sentence, with no explanations or extra text.\
• Use poetic or inverted word order when natural.\
• The style should resemble classical Shakespearean plays or sonnets.\

\medskip
Original: [sentence]
}
\end{quote}

\paragraph{(4) Authorship Style Transfer (Shakespearean $\rightarrow$ Modern)}
\begin{quote}
{\small\itshape
You are an expert in style transfer.\
Rewrite Shakespearean English into modern English while preserving the original meaning.\

\medskip
\textbf{Guidelines:}\
• Keep the same subject and meaning.\
• Output \textbf{only} the rewritten modern English sentence, with no explanations or extra text.\
• Simplify inverted or poetic word order into standard modern syntax.\
• Use clear, natural, contemporary English.\

\medskip
Original: [sentence]
}
\end{quote}

\section{Style Attributes Visualization}
\section{Case Study}

\begin{figure*}[t] 
    \centering
    \includegraphics[width=1\textwidth]{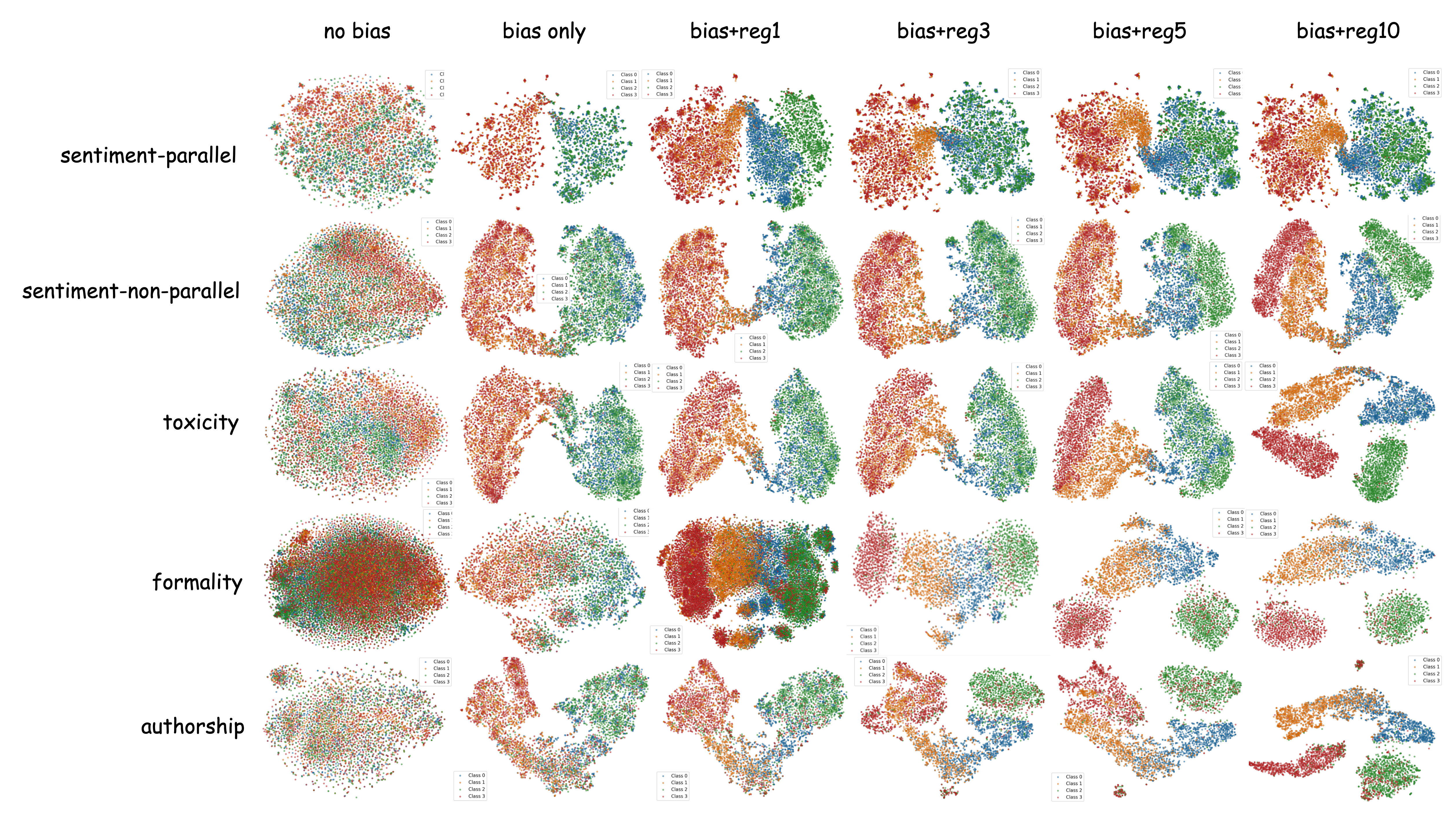} 
    \caption{This is the visualization of style attributes} 
    \label{fig:example} 
\end{figure*}
\clearpage

\end{document}